\documentclass[10pt,twocolumn,letterpaper]{article}

\usepackage[pagenumbers]{cvpr} 

\usepackage{graphicx}
\usepackage{amsmath}
\usepackage{amssymb}
\usepackage{booktabs}

\usepackage{lipsum}

\usepackage[pagebackref,breaklinks,colorlinks]{hyperref}

\usepackage{multirow}
\usepackage{float}

\usepackage{xcolor}

\usepackage[capitalize]{cleveref}
\crefname{section}{Sec.}{Secs.}
\Crefname{section}{Section}{Sections}
\Crefname{table}{Table}{Tables}
\crefname{table}{Tab.}{Tabs.}

\def\cvprPaperID{30} 
\def\confName{CVPR}
\def\confYear{2022}

\begin{document}

\title{FenceNet: Fine-grained Footwork Recognition in Fencing}

\author{
Kevin Zhu
\qquad
Alexander Wong
\qquad
John McPhee\\
University of Waterloo\\
{\tt\small \{k79zhu, a28wong, mcphee\}@uwaterloo.ca}
}
\maketitle

\begin{abstract}
Current data analysis for the Canadian Olympic fencing team is primarily done manually by coaches and analysts. Due to the highly repetitive, yet dynamic and subtle movements in fencing, manual data analysis can be inefficient and inaccurate. We propose FenceNet as a novel architecture to automate the classification of fine-grained footwork techniques in fencing. FenceNet takes 2D pose data as input and classifies actions using a skeleton-based action recognition approach that incorporates temporal convolutional networks to capture temporal information. We train and evaluate FenceNet on the Fencing Footwork Dataset (FFD), which contains 10 fencers performing 6 different footwork actions for 10-11 repetitions each (652 total videos). FenceNet achieves 85.4\% accuracy under 10-fold cross-validation, where each fencer is left out as the test set. This accuracy is within 1\% of the current state-of-the-art method, JLJA (86.3\%), which selects and fuses features engineered from skeleton data, depth videos, and inertial measurement units. BiFenceNet, a variant of FenceNet that captures the ``bidirectionality" of human movement through two separate networks, achieves 87.6\% accuracy, outperforming JLJA. Since neither FenceNet nor BiFenceNet requires data from wearable sensors, unlike JLJA, they could be directly applied to most fencing videos, using 2D pose data as input extracted from off-the-shelf 2D human pose estimators. In comparison to JLJA, our methods are also simpler as they do not require manual feature engineering, selection, or fusion.

\end{abstract}


\section{Introduction and background}
\label{sec:intro}
\begin{figure*}[t]
  \centering
  \includegraphics[width = 0.95\linewidth]{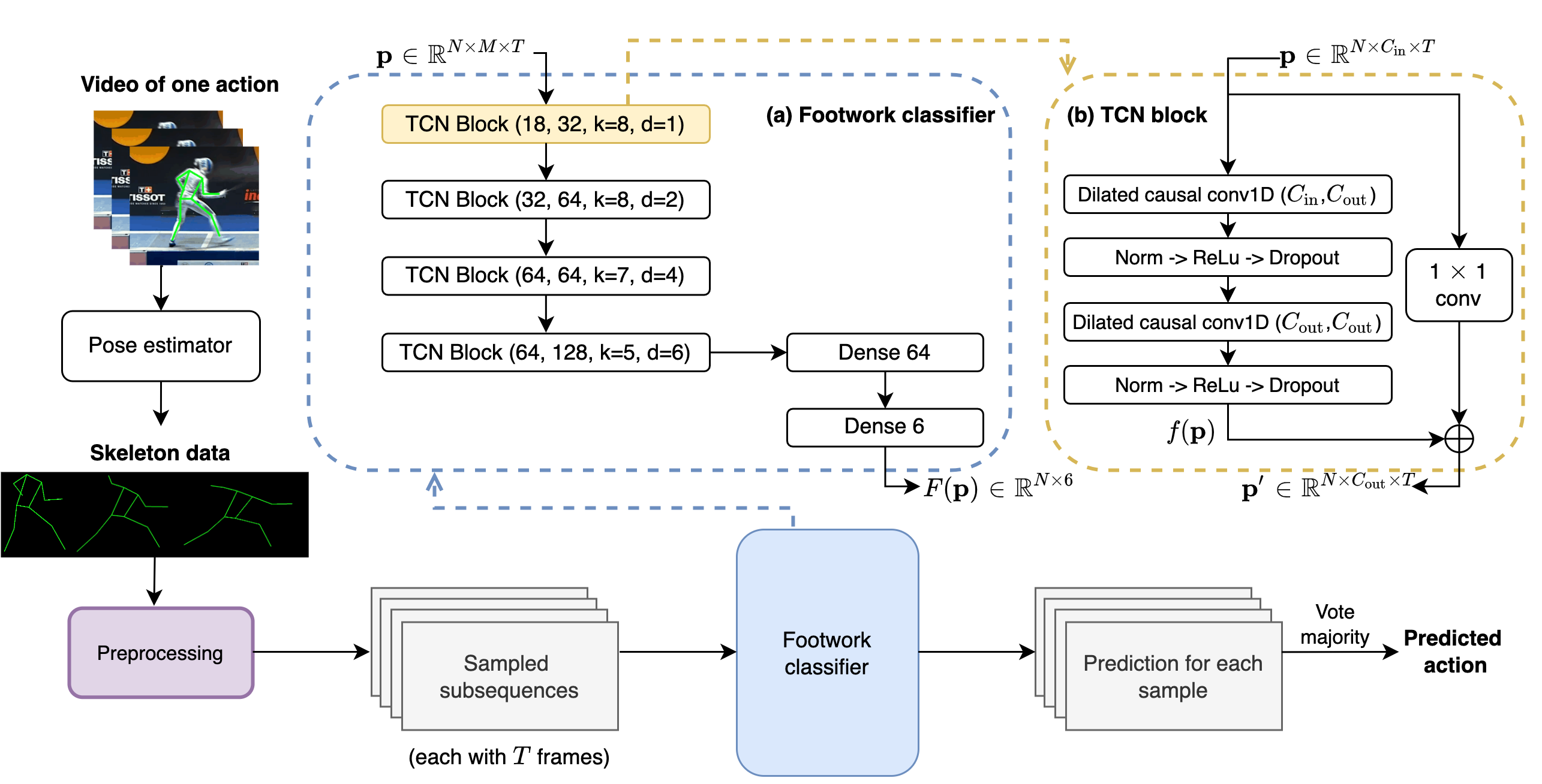}
  \caption{Network architecture of FenceNet. The footwork classifier (a) consists of stacked TCN blocks shown in (b). When training on 2D skeleton data from FFD, the pose estimator step is omitted.}
  \label{fig:framework}
\end{figure*}

There is a current need from national-level fencing teams for the development of analytical tools to enhance performance and training. The first step is to achieve a deeper understanding of the physical, tactical, and technical demands of fencing. Once these demands are better understood, performance benchmarks can be created for different skill levels to identify gaps and more accurately evaluate athletes. This in turn contributes to athlete selection, skills progression, and training interventions. The main bottleneck is the lack of a reproducible means to collect the high quality, high resolution, objective data that is required to create these benchmarks. 

Recognizing the need for automated analysis of techniques and motion in fencing, Malawski and Kwolek were among the first to apply computer vision approaches to detect and classify fine-grained actions in the sport. In \cite{Malawski2019, Malawski2018}, they proposed a method to classify fencing footwork by extracting 4 feature sets from visual and inertial signals, described below.

Joint dynamic (JD) features were proposed to describe the changes in motion of a fencer during the action, rather than the trajectory of motion. Skeleton data was split into windows of different sizes, for which the first 3 coefficients from the Short Time Fourier Transform computed for the velocity and acceleration of each joint along the 2 axes of 3 planes were used as features. Local trace image (LTI) features were proposed to represent the action as one image. Similar to motion energy images \cite{Bobick2001} and motion history images \cite{Davis1997}, a person's silhouette's binary images during the course of an action are superimposed, with a decay factor to better capture temporal information. To minimize noise, LTI crops out each joint, superimposes them separately, then resizes and concatenates them back together. Joint motion history context (JMHC) descriptors were proposed to capture local motion changes around joints. The absolute difference in silhouettes from depth images between two consecutive frames are described as histograms with each joint as the center. Histograms are normalized then concatenated to form a joint motion context (JMC) descriptor. The weighted sum of 3 consecutive JMC descriptors form a JMHC descriptor. Accelerometric (Acc) features from the time domain \cite{Parkka2006} were extracted from data captured by inertial measurement units (IMU).

Next, a feature selection algorithm based on feature ranking \cite{Chandrashekar2014} was proposed to reduce the dimensionality of each feature set. The reduced feature sets are then fused with a decision-level fusion scheme \cite{Mangai2010} by training a separate support vector machine (SVM) \cite{Cortes1995} for each feature set and concatenating the outputs, which are finally fed into a multilayer perceptron \cite{Rumelhart1986} for classification. We refer to this method as JLJA (JD+LTI+JMHC+Acc) moving forward.

JLJA was trained and evaluated on the Fencing Footwork Dataset (FFD)\cite{Malawski2018} that contains 6 basic fencing footwork actions -- stepping forward, stepping back, and 4 types of lunges with similar motion trajectories but subtle differences in dynamics. JLJA is currently the best performing method on FFD. However, the requirement of wearable sensors and depth video limits the pool of athletes to which JLJA can be applied.

To overcome this limitation, we propose a novel architecture, FenceNet, that takes only 2D skeleton data as input for the same classification task, and achieves similar accuracy to JLJA when evaluated on FFD. We also introduce a variant, BiFenceNet, that outperforms JLJA while using the same 2D skeleton data. This way, coaches and analysts could extract information directly from videos, by training FenceNet on 2D pose data extracted from an off-the-shelf 2D pose estimator \cite{Mcnally2021, Wang2021, Cao2021}, as seen in \cref{fig:framework}. FenceNet uses a skeleton-based human action recognition approach \cite{Liu2020, Choutas2018, Yan2018} that incorporates temporal convolutional networks (TCN) to capture temporal information.

The concept of a TCN was first introduced by Lea \etal \cite{Lea2017} for action segmentation and detection in videos. TCNs are mainly characterized by the use of two types of convolutions:
\begin{itemize}
    \item causal convolutions to ensure no leakage of future information into the current time step. 
    \item dilated convolutions to exponentially enlarge the receptive field.
\end{itemize}

\noindent Similar to recurrent neural networks (RNN) \cite{Jordan1986}, TCN models are able to take in a sequence of variable length and produce an output of the same length as the input. In comparison to RNNs, TCNs are generally faster and require less memory for training than RNNs. Since filters are shared across layers, convolutions are done in parallel, which allows TCNs to process the input sequence as a whole. On the other hand, RNNs process the input sequentially and often require more memory to store partial results. Performance-wise, empirical evaluations from Bai \etal \cite{Bai2018} showed that a TCN model was often able to achieve better results than RNN-based networks of similar size on various sequence modeling tasks.

FenceNet has the following advantages over JLJA:
\begin{itemize}
    \item \textbf{Transferability to competition videos.} Requiring only 2D skeleton data as input allows FenceNet to be transferred and trained on competition videos, including cases where access to additional data from wearable sensors and depth videos are unavailable. This allows coaches and analysts to extract information from fencers from other competition groups, other countries, and the past.
    
    \item \textbf{Transferability to other techniques.} Actions in fencing are highly composite. For example, an attack usually consists of a long sequence of varying movements used to counteract and react to the opponent's movements. JLJA splits feature vectors into windows of 16 frames, which limits memory retention. In contrast, due to dilated convolutions, TCNs have access to substantially longer memory, allowing FenceNet to be trained to classify other techniques in fencing.
    
    \item \textbf{Simplicity and automation.} Unlike JLJA, FenceNet does not require manual feature extraction, feature selection, or feature fusion.
\end{itemize}

\section{Related work}
\label{sec:related}
\subsection{Computer vision in fencing}

As described in \cref{sec:intro}, the majority of studies that involve computer vision applications in fencing were done by Malawski and Kwolek. In addition to their work on fencing footwork classification, they developed a model-based filtering algorithm for fencing footwork detection and segmentation on data acquired by a Kinect motion sensor \cite{Malawski2017}. In \cite{Malawski2018_2}, Malawski proposed a method for blade tracking based on a single RGB camera and active markers using augmented reality.

Earlier work includes the analysis of the lunge movement from video capture data \cite{Bober2016, Moore2015}. Mantovani \etal classified weapon actions on kinematic data acquired from a motion capture system \cite{Mantovani2010}.

More recent work includes the fencing tracking and visualization system developed from Rhizomatiks' collaboration with Dentsu Lab Tokyo \cite{rhizomatiks}. The system uses deep learning to detect sword tips without markers and real-time augmented reality synthesis to visualize the trajectory.

\subsection{Skeleton-based action recognition in sports}

Although end-to-end models\cite{Bertasius2021, Liu2021, Sudhakaran2020, Feichtenhofer2019, Lin2019, Tran2018, Carreira2017, wang2016} dominate the literature in video action recognition, they are often more suited for coarse-grained classification tasks. Classification in sports are generally more fine-grained \cite{Shao2020, Sun2017}. Being able to classify subtechniques, such as different types of punches, is often more useful for analysis than distinguishing between a punch and a kick. 
Skeleton-based methods have been a popular approach for fine-grained action recognition in sports. This method involves the use of 2D or 3D human pose as input in a human action recognition task.
Representing the human skeleton as a graph with joint positions as nodes and modeling movement as the change of these graph coordinates over time allows us to capture both the spatial and temporal components of the action. Non-deep learning based approaches have been explored in sports such as wrestling \cite{Mottaghi2020} and Tai Chi \cite{Xu2020}.

In addition to improved performance, deep learning offers many advantages, such as automating feature engineering and feature selection. RNNs were one of the first networks used to model the temporal component of human actions. Long short-term memory (LSTM) \cite{Hochreiter1997} based RNNs specifically, were commonly used because traditional RNNs suffer from the vanishing/exploding gradient problem \cite{Pascanu2013, Hochreiter1997}. Variants that incorporate graph convolutional networks (GCNs) such as GT-LSTM \cite{Li2020} and LSGM \cite{Huang2020} were proposed to capture spatial information for skeleton-based action recognition.

More recently, TCN architectures have been used in place of RNN-based layers due to their ability to exhibit longer memory \cite{Bai2018}. In table tennis, Kulkarni and Shenoy \cite{Kulkarni2021} used TCNs for stroke prediction and showed that their TCN model outperformed their LSTM model.

\section{Fencing videos}
\label{sec:data}
FenceNet is trained on FFD, a publicly available fencing dataset that contains 10 intermediate to expert level fencers performing 6 types of footwork actions (lunges and steps) for 10-11 repetitions each in a practice setting, for a total of 652 videos. Czajkowski \cite{Czajkowski2005}, known as one of the inventors of modern fencing theory, grouped the fencing lunge into 4 categories. The 6 total actions, with descriptions of the lunges by Czajkowski, are:

\begin{itemize}
    \item \textit{rapid lunge (R):} very fast, performed in relatively short distances.
    \item \textit{incremental speed lunge (IS):} slow at beginning, accelerates during action, useful for feint attacks.
    \item \textit{with waiting lunge (WW):} short pause in first stage of lunge  while fencer observes reaction of opponent to counter-action.
    \item \textit{jumping sliding lunge (JS):} fencer jumps forward with front leg to cover distance, back leg slides on the floor, common in complex offensive actions.
    \item \textit{step forward (SF)}
    \item \textit{step backward (SB)}
\end{itemize}

\begin{figure}[H]
  \centering
  \includegraphics[width = 0.3\linewidth]{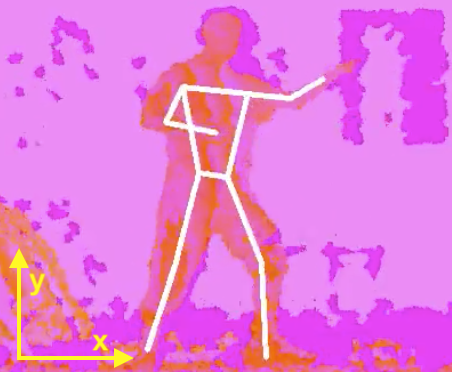}
  \includegraphics[width = 0.3\linewidth]{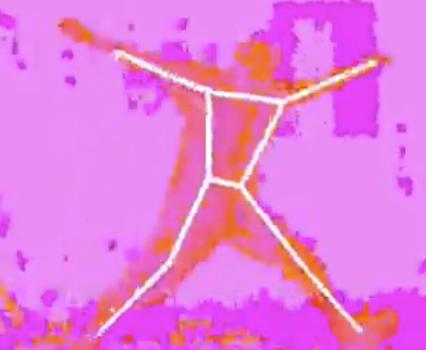}
  \includegraphics[width = 0.3\linewidth]{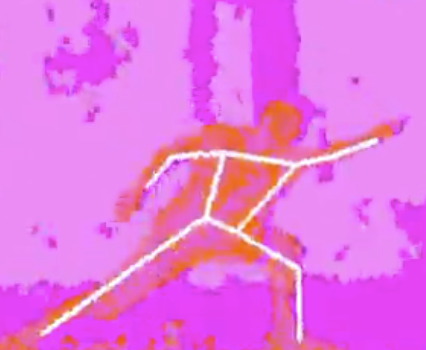}
  \caption{Skeleton data overlayed on depth data from FFD. Only 2D skeleton data was used for this study.}
  \label{fig:ffd}
\end{figure}

FFD contains 3D skeleton data and 640$\times$480 16-bit depth data acquired by Kinect \cite{kinect} at 30 Hz. 9 axis accelerometer, gyroscope, magnetometer, and orientation data were captured by an x-IMU sensor at 256 Hz. For this study, we only use the $x,y$ coordinates (see \cref{fig:ffd}) of the skeleton data.

When processing the FFD files, we noticed that the skeleton data for the second repetition of \textit{SF} for fencer 5\footnote{File name: \textit{2016-01-09\_12-51-53.Body.mat}} was empty and thus removed for this study.

\section{FenceNet}
\label{sec:method}
\subsection{Preprocessing phase}
\label{dataprocess}
From \cref{tab:framecount} we see that different actions have different frame counts, with lunges being longer than steps. To allow for batch training, we sample 28 consecutive frames from each video with a random starting point from the beginning to the 20th frame. Each video is sampled at most 10 times (videos with fewer than 47 frames had less than 10 samples). We chose a window size of 28 since that is the minimum frame count for all videos. 

\begin{table}[H]
\centering
\begin{tabular}{lccccccc}
\hline
  &                                                                              mean & std & min  & 25\%  & 50\% & 75\% & max  \\ \hline
  R                                                                         & 53.5 & 5.7 & 40 & 50 & 53 & 57 & 68 \\
IS                                                                       & 65.1 & 8.8 & 49 & 58 & 64 & 72 & 98 \\
WW                                                                        & 70.4 & 8.7 & 52 & 64 & 70 & 76 & 92 \\
JS                                                                       & 69.9 & 9 & 51 & 62 & 70 & 75 & 98 \\

SB                                                                        & 41.1 & 8.1 & 28 & 33 & 41 & 48 & 62 \\
SF                                                                       & 44.2 & 9.8 & 29 & 37 & 42 & 50 & 80 \\
 \hline
\end{tabular}
\caption{Summary of frame counts for each action in FFD.}
\label{tab:framecount}
\end{table}


Since \textit{SF} and \textit{SB} had significantly fewer frames per video than the lunge actions, sampling inevitably introduced some class imbalance to our data, as seen in \cref{tab:videocounts}. However, since the steps are the coarse actions, while the differences in motion among lunges are subtle, this actually helped our fine-grained action recognition performance via a form of data augmentation. This is discussed more in \cref{sec:expSampling}.


\begin{table}[H]
\centering
\begin{tabular}{lcc}
\hline
   & Before sampling & After sampling \\ \hline
R  & 108 (16.7\%)    & 1053 (18.3\%)  \\
IS & 110 (16.8\%)    & 1100 (19.1\%)  \\
WW & 110 (16.8\%)    & 1100 (19.1\%)  \\
JS & 109 (16.7\%)    & 1090 (18.9\%)  \\
SF & 107 (16.4\%)    & 761 (13.2\%)   \\
SB & 108 (16.5\%)    & 660 (11.5\%)   \\ \hline 
Total & 652    & 5764    \\\hline
\end{tabular}
\caption{Video counts for each action before and after sampling.}
\label{tab:videocounts}
\end{table}

For each sampled subsequence, we subtract the fencer's nose's position of the first frame from every joint coordinate in each frame. Then each joint coordinate in each frame is divided by the vertical distance between the head position and front ankle in the first frame. Letting $p^{j,c}_t$ be the position of joint $j$ on the $c$ axis during time step $t$, the scaled position $\tilde{p}^{j,c}_t$ is given by:

\begin{align}
    \tilde{p}^{j,c}_t = \frac{p^{j,c}_t - p^{N,c}_0}{p^{N,c}_0 - p^{A,c}_0}
\end{align}
\noindent where $N$ and $A$ represent keypoints for the nose and front ankle, $0< t\leq 28$, and $c\in\{x,y\}$.

We take the $x,y$ coordinates of the front wrist, front elbow, front shoulder, both hips, both knees, and both ankles as inputs to our classification model.


\subsection{Footwork classifier}

\begin{figure}[H]
  \centering
  \includegraphics[width = 0.95\linewidth]{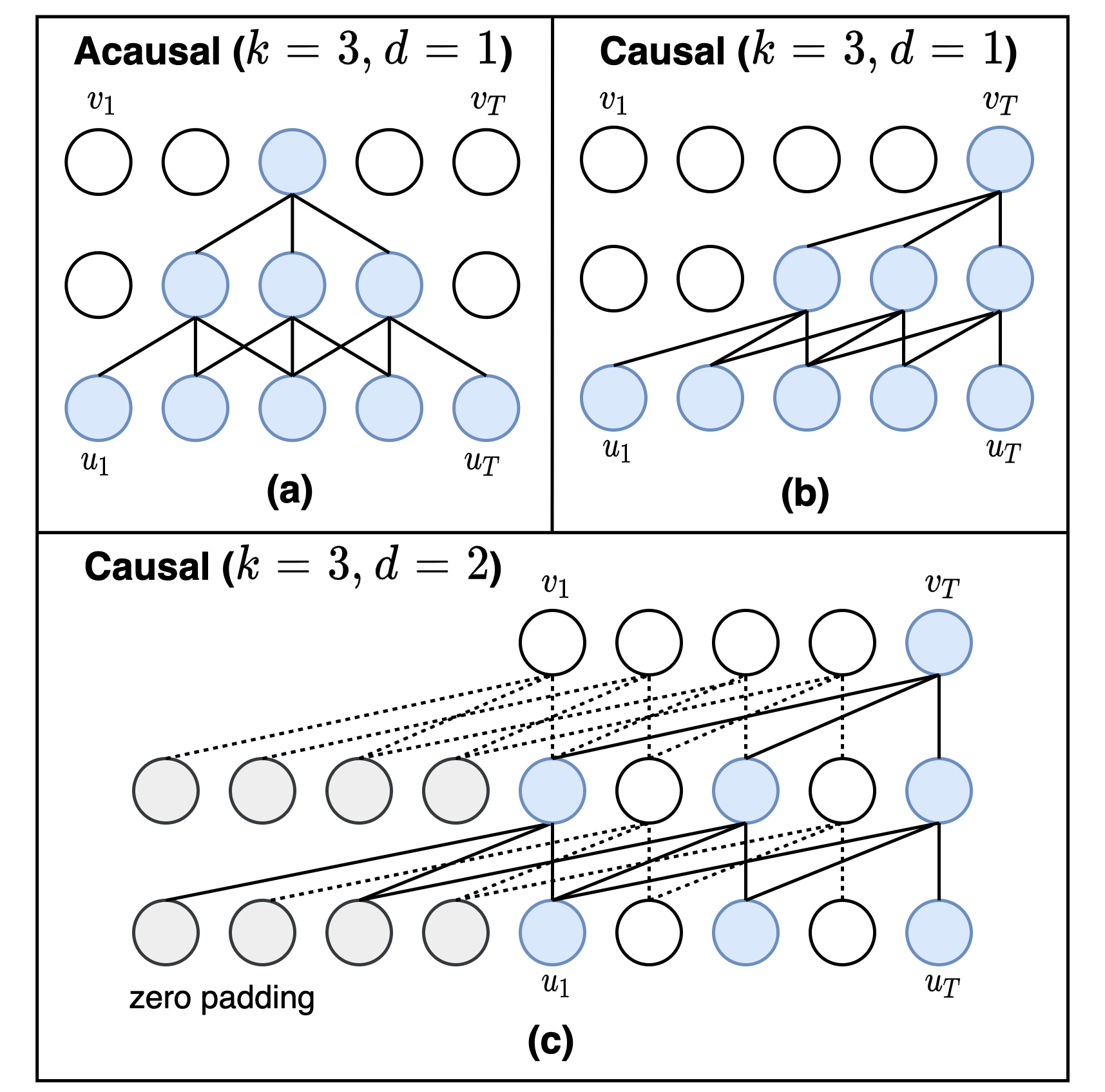}
  \caption{Given input sequence $\mathbf{u}$ and output sequence $\mathbf{v}$ (both of length $T=5$),  kernel size $k=3$, and one hidden layer, an example of (a) normal convolutions. (b) causal convolutions. (c) dilated causal convolutions with dilation factor $d=2$ and zero padding to ensure same length in each layer. Note: some connections in (a) and (b) are not drawn.}
  \label{fig:conv}
\end{figure}

FenceNet consists of 6 TCN blocks followed by two dense layers (see \cref{fig:framework}a). A TCN block (\cref{fig:framework}b) contains two stacked 1D fully-convolutional layers \cite{Long2015}, each employing causal convolutions and dilated convolutions \cite{Yu2016}. Causal convolutions (\cref{fig:conv}b) ensure there is no leakage of information from the future to the past, meaning predictions made at time $t$ depend only on states during and prior to $t$. To achieve this, for time step $t$ and kernel size $k$, we convolve from $t-k$ to $t$ (as opposed to from $t-\frac{k}{2}$ to $t+\frac{k}{2}$ in the acausal case). Let $C_{causal}(\mathbf{p}, t)$ denote causal convolution at step $t$ for input $\mathbf{p}$:

\begin{align}
    C_{causal}(\mathbf{p}, t) = \sum_{i=0}^{k-1}w(i)\ast\mathbf{x}_{t-i}
\end{align}
\noindent where $\ast$ is the cross-correlation operator and $w$ is the filter.

Dilated convolutions allow us to exponentially increase receptive field size in different ways, such as by increasing the number of dilated convolutional layers, increasing the dilation factor, or kernel size. In contrast, the receptive field of regular convolution is linear to depth or kernel size (\cref{fig:conv}a). Thus dilated convolutions provide better control over model size and complexity which can reduce the risk of overfitting. Furthermore, dilated convolutions could be used in place of pooling and upsampling to better retain information between layers. Dilated convolutions are implemented by skipping a fixed gap between time steps. For example, given dilation factor $d$ and kernel size $k$, when combined with causal convolutions (\cref{fig:conv}c), at step $t$ we have:
\begin{align}
    C_{dilated\; causal}(\mathbf{p}, t, d) = \sum_{i=0}^{k-1}w(i)\ast\mathbf{p}_{t-d\cdot i}
\end{align}

The convolutional layers are immediately followed by weight normalization \cite{Salimans2016}, a rectified linear unit (ReLU) \cite{Nair2010}, and spatial dropout \cite{Srivastava2014} to improve generalization. Lastly a residual connection is added between the input and output of each block to improve stability of the network. Given input $\mathbf{p}$ and output $f(\mathbf{p})$, this residual connection is simply:
\begin{align}
    \text{output} = \text{Activation}(\mathbf{p} + f(\mathbf{p}))
\end{align}

In the case that the input channel size could differ from the output channel size of the second convolutional layer, a 1$\times$1 convolution is added to account for this discrepancy. Our structure is based on those used in \cite{Luo2021, Bai2018} but with increasing hidden size and decreasing kernel size as the number of layers increase. The increasing dilation factors are also adjusted to accommodate for the limited input length while maintaining full history coverage. To ensure each layer has the same length, zero padding is used.

From the output sequence of the last TCN block, we extract the last time-step and feed it into dense layers for prediction. Due to sampling in \cref{dataprocess}, for a given video, we have a predicted action for each subsequence of frames. We select the most commonly predicted action among the subsequences as our final predicted action. Details of the structure and parameters of the network can be found in \cref{fig:framework}a. These values were tuned using random search.


\subsection{BiFenceNet}
\label{causal}
\begin{figure}[H]
  \centering
  \includegraphics[width = 0.95\linewidth]{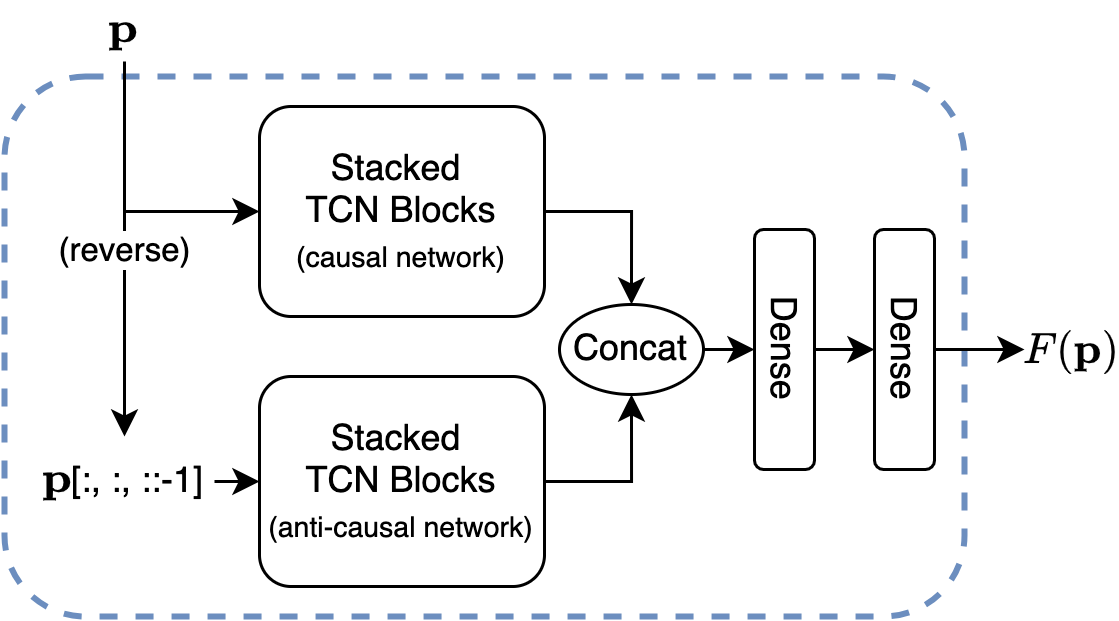}
  \caption{A bidirectional TCN-based module that replaces the footwork classifier (\cref{fig:framework}a) in BiFenceNet. The TCN blocks remain the same as in \cref{fig:framework}b.}
  \label{fig:bfn}
\end{figure}

Causal convolutions ensure no information leakage into the future, which allows us to sequentially capture the forward motion of an action. Inspired by ELMo \cite{Peters2018_elmo}, we hope to capture ``bidirectionality'' of motion by using two separate networks. As seen in \cref{fig:bfn}, we capture the forward motion of an action through a network of stacked TCN blocks, while feeding the reversed motion into a separate stack of TCN blocks, essentially creating a separate ``anti-causal" network. The TCN blocks in the two networks are the same as in \cref{fig:framework}b. As in FenceNet, we extract the last time step from the output of each network. They are concatenated and fed into dense layers for prediction. To create BiFenceNet, we simply replace the footwork classifier seen in \cref{fig:framework}a with the bidirectional TCN-based module in \cref{fig:bfn}.

\section{Experimental results}
\label{sec:experiments}
\subsection{Evaluation}
Our model is trained and evaluated on FFD using 10-fold cross-validation, where in each fold, data from one fencer is taken out as the test set. This scenario provides a better representation of how the model generalizes to new fencers than randomly splitting the training and testing data. Malawski and Kwolek referred to this scenario as the person-independent (PI) case when evaluating JLJA. Since Malawski and Kwolek did not provide the train-test split for their random 5-fold cross-validation scenario, we omit the comparison for that case. The JLJA result displayed in this section is the top performing result from the various combinations of features and parameters used in \cite{Malawski2019, Malawski2018}. 

Under the PI case, FenceNet achieved a classification accuracy of 85.4\%, within 1\% of JLJA (86.3\%), after training for 103 epochs for all 10 folds. BiFenceNet achieved a classification accuracy of 87.6\%, outperforming JLJA after training for 94 epochs for all 10 folds, with 4 layers for each of the two stacked TCN blocks. In addition to JLJA, Malawski and Kwolek also evaluated methods such as SkeletonNet \cite{Ke2017}, C3D \cite{Tran2015}, EigenJoints \cite{Yang2014}, HON4D \cite{Oreifej2013}, LOP/FTP \cite{Wang2012}, and MHI \cite{Bobick2001}. Comparisons of the methods can be found in \cref{tab:resall} (note that results from all non-FenceNet methods were computed by Malawski and Kwolek in \cite{Malawski2019}). As mentioned in \cref{sec:data}, we removed 1 of the 653 files from FFD for this study as we were unable to process the data in that file. However, since the removed file was an \textit{SF} action, and we were able to separate this class well in both FenceNet and BiFenceNet (\cref{tab:cm_fn} and \cref{tab:cm_bfn}), we believe this one missing observation will not alter our results and that our results are still comparable to the other methods shown in \cref{tab:resall}.

\begin{table}[H]
\centering
\begin{tabular}{lc}
\hline
 Method            & Accuracy \% \\ \hline
JLJA \cite{Malawski2019} *        & \textbf{86.3}                                                                                                       \\
EigenJoints \cite{Yang2014} * &      29.9                                                                                                  \\ 
MHI \cite{Bobick2001} * &         61.3                                                                                               \\ 
SkeletonNet \cite{Ke2017} * &      64.4                                                                                             \\ 
C3D \cite{Tran2015} * &        67.6                                                                                                \\ 
HON4D \cite{Oreifej2013} * &         75.9                                                                                          \\ 
LOP/FTP \cite{Wang2012} * &      76.1          \\ \hline
FenceNet (ours)         & \textbf{85.4}                                                                                                      \\
BiFenceNet (ours)       & \textbf{87.6}                                                                                                     \\
\hline
\end{tabular}
\caption{Classification accuracy for the PI case (* results taken directly from Sec. 5 of \cite{Malawski2019}).}
\label{tab:resall}
\end{table}


From the confusion matrices (Tabs. \ref{tab:cm_fn}, \ref{tab:cm_bfn}, \ref{tab:cm_jlja}) we observe that, compared to JLJA, our methods are also better at distinguishing between coarse actions -- steps from lunges. This could be useful in future action segmentation tasks for fencing matches.

For all three methods (Tabs. \ref{tab:cm_fn}, \ref{tab:cm_bfn}, \ref{tab:cm_jlja}), we find \textit{IS} to be the worse performing class, often mixed with \textit{WW}. This is likely due to different fencers subjectively interpreting the two classes differently, as the two actions share almost identical motion trajectories but differ in speeds at different time points. Having some prior knowledge of the fencer can improve results. For example, during the random split case, where we randomly take out 20\% of the repetitions for each action for each fencer as the test set, FenceNet achieves a significantly higher accuracy of 96.5\% on the test set.
JLJA's ability to better identify \textit{IS} is likely due to its JMHC descriptor directly incorporating the change in depth image between consecutive frames to better capture the ``incremental" change in speed of the movement, whereas in FenceNet, only skeleton data is used as input. This can cause FenceNet to misclassify ``faster" \textit{IS} repetitions as \textit{R} and ``slower" ones as \textit{WW}, since different athletes perform techniques at different overall speeds. Future variants of FenceNet can explore incorporating the change in skeleton data as input.
Furthermore, distinguishing \textit{IS} and \textit{WW} in a binary classification scenario could be a focus in future work. A hierarchical approach where \textit{IS} and \textit{WW} are treated as one class during initial classification and separated in the second stage as a binary case may improve overall accuracy.


\begin{table}[H]
\centering
\begin{tabular}{lllllll}
\hline
   & R               & IS              & WW            & JS            & SF   & SB   \\ \hline
R  & \textbf{88.9}   & 7.4            & 1.9             & 1.9          & -    & -    \\
IS & 15.5             & \textbf{51.8}   & 17.3             & 15.5         & -    & -    \\
WW & 0.9             & 15.5            & \textbf{82.7} &  0.9            & -    & - \\
JS & -             & 10.1             & -            & \textbf{89.9} & -    & -    \\
SF & -               & -               & -           & -             & \textbf{100} & - \\
SB & -               & -               & -             & -             & - & \textbf{100} \\ \hline
\end{tabular}
\caption{Confusion matrix for FenceNet for the PI case (prediction accuracy \textbf{85.4\%}).}
\label{tab:cm_fn}
\end{table}

\begin{table}[H]
\centering
\begin{tabular}{lllllll}
\hline
   & R               & IS              & WW            & JS            & SF   & SB   \\ \hline
R  & \textbf{95.4}   & 0.9            & -            & 3.7          & -    & -    \\
IS & 14.5             & \textbf{59.1}   & 13.6             & 12.7         & -    & -    \\
WW & 1.8             & 14.5            & \textbf{83.6} & -            & -    & - \\
JS & -             & 3.7            & 7.3             & \textbf{89.0} & -    & -    \\
SF & -               & -               & 0.9           & -             & \textbf{99.1} & - \\
SB & -               & -               & -             & -             & - & \textbf{100} \\ \hline
\end{tabular}
\caption{Confusion matrix for BiFenceNet for the PI case (prediction accuracy \textbf{87.6\%}).}
\label{tab:cm_bfn}
\end{table}

\begin{table}[H]
\centering
\begin{tabular}{lllllll}
\hline
   & R             & IS            & WW            & JS            & SF   & SB   \\ \hline
R  & \textbf{85.3} & 12.0          & 1.8          & 0.9          & -    & -    \\
IS & 11.0          & \textbf{71.6} & 5.6           & 11.8         & -    & -    \\
WW & 4.6           & 18.2          & \textbf{77.3} & -         & -    & - \\
JS & -          & 13.6          & -             & \textbf{86.4} & -    & -    \\
SF & -          & -             & -             & -             & \textbf{100} & - \\
SB & -             & -             & -             & -             & 2.7 & \textbf{97.3} \\ \hline
\end{tabular}
\caption{Confusion matrix for JLJA for the PI case (prediction accuracy \textbf{86.3\%}), taken from the top performing version in \cite{Malawski2019}.}
\label{tab:cm_jlja}
\end{table}


\begin{table*}[t]
\centering
\begin{tabular}{lccccccccc}
\hline
                         & \multirow{2}{*}{\begin{tabular}[c]{@{}c@{}}Parameters \\ (10$^6$)\end{tabular}} 
                         & \multicolumn{1}{c}{\multirow{2}{*}{\begin{tabular}[c]{@{}c@{}}Prediction \\ accuracy(\%)\end{tabular}}} 
                         & \multicolumn{6}{c}{Class accuracies (\%)} \\
                         & & 
                         \multicolumn{1}{c}{} & R & IS & WW & JS & SF & SB \\ \hline

FenceNet    & 2.6   & \textbf{85.4} & 89   & 52   & 83   & 90   & 100   & 100   \\
BiFenceNet    & 5.4  & \textbf{87.6} & 95   & 59   & 84   & 89   & 99   & 100   \\ 
\hline
FenceNet (reversed)   & 2.6  & 84.4 & 86   & 61   & 78   & 83   & 99    & 100   \\

FenceNet (shuffled) & 2.6 & 84.4 & 87   & 50   & 84   & 87   & 100    & 99   \\

FenceNet (forward $\times$2)   & 7.0    & 86.2 & 95 & 50   & 86 & 86 & 100 & 100   \\ 

FenceNet (wide)  & 5.9  & 85.4  & 90 & 56 & 80 & 88 & 100 & 100  \\ 

FenceNet (regular conv1D)   & 4.8    & 83.3 & 92 & 41 & 83 & 84 & 100 & 100   \\ 
\hline
FenceNet (zero padding)   &  2.6  &   76.5  & 82  & 28 & 74 & 77 & 100  & 100    \\

FenceNet (full body)   & 2.7 & 83.1 & 89   & 52   & 72   & 87   & 100   & 100   \\

FenceNet (lower body)  & 2.6 & 82.4 & 77   & 60   & 72   & 90   & 99    & 97   \\ 
\hline
LSTM  & 2.7    & 81.9   & 92 & 34 & 78 & 89 & 100 & 100 \\ 

Bi-LSTM  & 5.5 & 83.1   & 93 & 39 & 80 & 88 & 100 & 100 \\ 

\hline
\end{tabular}
\caption{Experimental results. Rows 3-7 correspond to results in \cref{sec:expCaus}. Row 8 corresponds to \cref{sec:expSampling}. Rows 9-10 correspond to \cref{sec:expKeypoint}. Rows 11-12 correspond to \cref{sec:expModel}. All methods were evaluated under the PI case.
}
\label{tab:expall2}
\end{table*}

\subsection{Causality}
\label{sec:expCaus}
To investigate the effect of causality, we examine cases where the input sequence is reversed (anti-causal) and shuffled (acausal). FenceNet outperforming these cases (rows 3-4 of \cref{tab:expall2}) provides evidence that the forward trajectory of motion contains useful information when distinguishing actions.

To investigate the effect of the additional network in BiFenceNet that aims to capture the reverse direction of movement, we replace the anti-causal network with another causal network. BiFenceNet outperforming this case (\cref{tab:expall2} row 5) provides evidence that the reversed motion trajectory contains additional information for distinguishing actions, and that the better performance from BiFenceNet is not simply due to an ensemble effect. Different parameters and structures for the second causal network were tested and the best result was recorded. In \cref{tab:expall2} row 6, we compare BiFenceNet to a wider version of FenceNet by increasing channel size to show that the improved performance is not simply due to an increase in model size. In \cref{tab:expall2} row 7, we compare BiFenceNet to a version of FenceNet with only dilated 1D convolution layers and no causality to show that training the forward and reverse directions of the input sequence separately is different from using regular 1D convolutions. Instead of taking the last time step of the output from the last block, we flatten the last 1D convolution layer before feeding into the dense layers. 


\subsection{Sampling versus zero padding}
\label{sec:expSampling}
During the preprocessing phase (\cref{dataprocess}), to ensure videos of the same length for batch training, an alternative to sampling would be to pad zeros to the end of shorter videos, as done in \cite{Kulkarni2021}. However, we chose to sample subsequences as this process simultaneously augments our training data.

Segmenting a sequence of movements into actions, even by manual cropping, is prone to error, and could lead to inconsistencies in defining the start and end of an action. Sampling subsequences of frames essentially augments the training data, and allows the model to better deal with this problem. From \cref{tab:expall2} row 8, we see that FenceNet outperforms the zero padding case significantly.


\subsection{Keypoint selection}
\label{sec:expKeypoint}
Despite the lunge being characterized as a lower body movement, we included the front wrist, elbow, and shoulder joint into our input. This is because Czajkowski described \textit{IS} as often being associated with feint attacks and \textit{WW} with counter-actions. We hypothesize that information extracted from the front arm could capture some of the aforementioned associations, and improve our lunge prediction. We compare this to using keypoints from the whole body (by including the nose and back arm), as well as only the lower body (both hips, both knees, both ankles). From rows 9-10 of \cref{tab:expall2}, we see that both these cases perform worse than our original method, aligning with our hypothesis.


\subsection{Model comparison}
\label{sec:expModel}
Lastly, \cref{tab:expall2} row 11 shows FenceNet outperforming an LSTM of similar size and  \cref{tab:expall2} row 12 shows BiFenceNet outperforming a bidirectional LSTM of similar size. The training times for the LSTM and bidirectional LSTM were both more than 3 times that of FenceNet and BiFenceNet, respectively. These results align with observations from \cite{Bai2018, Kulkarni2021}, which state that TCNs often outperform LSTMs in sequential tasks.

\section{Discussion}
\label{sec:results}
A limitation of FenceNet is its dependency on the quality of 2D pose input. Without a marker-based motion capture system, pose data extracted from off-the-shelf 2D pose estimators could be noisy or inaccurate, limiting the performance of FenceNet. To address this issue, future work can focus on the preprocessing step to obtain more robust inputs. This includes data augmentation, smoothing, pose normalization, or incorporating methods such as VIPE \cite{Sun2020} to directly extract pose features from 2D input.

The next step involves collecting high quality labeled fencing competition video data to test these methods to obtain more robust results. Competition video data also allows us to further explore different computer vision tasks in fencing, such as action segmentation and retrieval of more complex techniques.

Preliminary data collection and tagging on competition videos are currently being conducted with help from video analysts from the Canadian Olympic fencing team. We found actions from FFD not directly transferable to a competition setting. The footwork classes are highly imbalanced. For example, in the Grand Prix Turin 2020 Women's Foil Final\footnote{\url{https://www.youtube.com/watch?v=H-v6DfxnjF8}}, we tagged all 45 lunges (2266 total frames) executed by French fencer Ysarora Thibus according to the description in \cref{sec:data} and noticed that the frequency of the \textit{JS} lunge dominates the other actions, as seen in \cref{tab:comp}.

\begin{table}[H]
\centering
\begin{tabular}{lc}
\hline
Action & Count \\ \hline
Rapid lunge (R)     & 3 (6.7\%)     \\
Incremental speed (IS)     & 4 (8.9\%)    \\
With waiting (WW)     & 3 (6.7\%)      \\
Jumping sliding (JS)     & 35 (77.8\%)   \\ \hline
\end{tabular}
\caption{Class frequency for each lunge performed by Ysarora Thibus in the Grand Prix Turin 2020 Women's Foil Final. The lunges are tagged according to the descriptions in \cref{sec:data}.}
\label{tab:comp}
\end{table}

Although there are multiple scenarios where Thibus adjusts the speed of her lunge as in \textit{IS} or \textit{WW}, they are almost always accompanied by a jumping, sliding motion in the lower body, as described by \textit{JS}. She very rarely keeps her back foot still during the entirety of a lunge, which is the case for all non-\textit{JS} lunges in FFD. For footwork recognition in a competition setting, we recommend creating more classes or modifying the definitions of the current ones, as techniques performed during competition seem to be much more complex and dynamic.

\section{Conclusion}
\label{sec:conclusion}
The current state-of-the-art fencing footwork recognition algorithm, JLJA, fuses information from skeleton data, depth videos, and IMU sensors to classify footwork techniques in fencing. However, the requirement of depth videos and wearable sensors imposes technological and physical difficulties on users, limiting the pool of athletes that they are able to analyze. To address this shortcoming, we introduce FenceNet, a fencing footwork recognition model that relies only on 2D pose data. FenceNet is a lightweight network that incorporates skeleton-based human action recognition with TCN architectures to capture both spatial and temporal components of human motions. Experimental results indicate that FenceNet's classification accuracy came within 1\% of JLJA, while BiFenceNet was able to outperform JLJA, despite both only using 2D skeleton data as input. We hope FenceNet and future variants are able to contribute to automating the analysis in sports.\\

\noindent\textbf{Acknowledgements.} We acknowledge financial support from the Canada Research Chairs Program, the Canadian Sports Institute Ontario, and a Mitacs grant. We also acknowledge Fencing Canada for their help with data collection, tagging, and fencing expertise.











{
\small
\bibliographystyle{ieee_fullname}
\bibliography{ref}
}

\end{document}